# Interpretability from a new lens: Integrating Stratification and Domain knowledge for Biomedical Applications

Anthony Onoja[1], Francesco Raimondi[2]
(a.onoja@surrey.ac.uk[1], francesco.raimondi@sns.it[2])
[1]School of Health Sciences, Faculty of Health and Medical Sciences, University of Surrey, United Kingdom,
[2]Laboratorio di Biologia Bio@SNS, Scuola Normale Superiore, Pisa, Italy

**Abstract**
*The use of machine learning (ML) techniques in the biomedical field has become increasingly important, particularly with the large amounts of data generated by the aftermath of the COVID-19 pandemic. However, due to the complex nature of biomedical datasets and the use of black-box ML models, a lack of trust and adoption by domain experts can arise. In response, interpretable ML (IML) approaches have been developed, but the curse of dimensionality in biomedical datasets can lead to model instability. This paper proposes a novel computational strategy for the stratification of biomedical problem datasets into k-fold cross-validation (CVs) and integrating domain knowledge interpretation techniques embedded into the current state-of-the-art IML frameworks. This approach can improve model stability, establish trust, and provide explanations for outcomes generated by trained IML models. Specifically, the model outcome, such as aggregated feature weight importance, can be linked to further domain knowledge interpretations using techniques like pathway functional enrichment, drug targeting, and repurposing databases. Additionally, involving end-users and clinicians in focus group discussions before and after the choice of IML framework can help guide testable hypotheses, improve performance metrics, and build trustworthy and usable IML solutions in the biomedical field. Overall, this study highlights the potential of combining advanced computational techniques with domain knowledge interpretation to enhance the effectiveness of IML solutions in the context of complex biomedical datasets.*

## 1. Introduction

Machine Learning (ML) is a branch of artificial intelligence (AI) that enables computers to gain knowledge and make predictions or decisions based on data, without requiring explicit programming [1]. It involves training statistical models on large datasets to recognize patterns and predict outcomes. ML is widely used in various fields such as finance, healthcare, and marketing. There are several types of ML techniques such as supervised learning, unsupervised learning, and reinforcement learning [2]. Supervised learning is the most common type, where the algorithm is trained on labeled data and uses this information to make predictions on new, unseen data. Unsupervised learning is a type of ML where the algorithm is not given any labeled data and must find patterns or structures in the data on its own [3], [4]. Reinforcement learning is a type of ML where the algorithm learns through trial-and-error interactions with its environment. However, ML models can be complex and difficult to understand, especially when they involve deep neural networks (DNN) [5]–[7]. These models are often referred to as "black box" models because it is difficult to understand how they arrived at their decisions. Complexity in the type of ML algorithm employed is related to the data type from which patterns and new insights experts sought to investigate [8]. Complexities in problem data are steadily increasing due to the veracity of data types collected from different big data sources (textual, images, etc.,). Recently, the biomedical domain has been transformed and witnessed an increase in different data types [9]–[11]. These data types hold a lot of information that can be harnessed to assist existing medical procedures [12]. However, the existing traditional approaches such as Statistics are limited in addressing some of these issues [13], [14]. The rise in ML techniques has provided a useful computational and efficient platform to harness useful information from these data types [5], [15]. The purpose of this paper is to propose a novel computational strategy for the integration of domain knowledge interpretation techniques into current state-of-the-art IML frameworks to improve the stability of IML models, establish trust and provide explanations for outcomes generated by trained IML models, and ultimately enhance the effectiveness of IML solutions in the context of complex biomedical datasets. This approach involves the stratification of biomedical problem datasets into k-fold (training and testing sets) cross-validation (CVs) and linking the model outcome, such as aggregated feature weight importance across the k-fold CVs, to further domain knowledge interpretations using techniques like pathway functional enrichment, drug targeting, and repurposing databases. This paper also emphasizes the importance of involving end-users and clinicians in focus group discussions to guide testable hypotheses, improve performance metrics, and build trustworthy and usable IML solutions.

The rest of this paper is organized as follows: section 2 discussed the application of ML techniques in biomedicals, section 3 discussed the interpretable ML approaches in Biomedicals, section 4 highlighted the need for splitting the problem dataset into several k-fold CVs via embedding a novel computational splitting strategy and the integration of domain knowledge approaches into the current state-of-the-art IML. lastly, the paper proposed a novel IML framework for biomedical data science problems. Section 5 concluded the study.

Figure 1 showed the summary of the current state-of-the-art ML framework used in a typical data science project lifecycle [16].

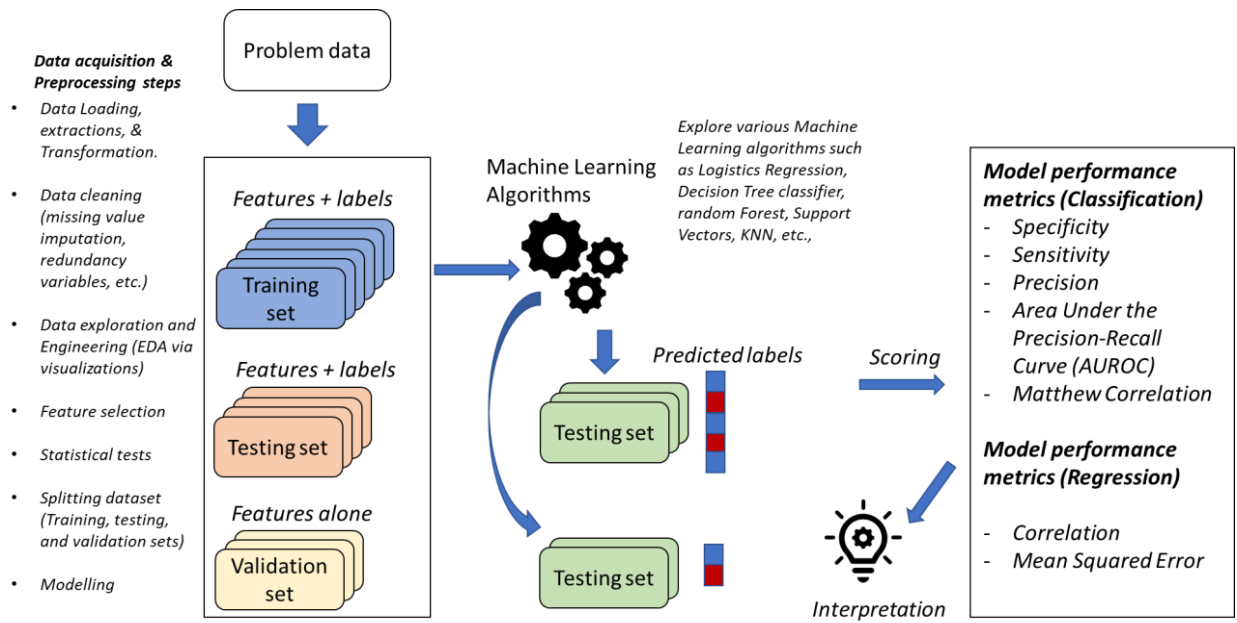

**Figure 1:** Current state-of-the-art Machine Learning framework in Data Science project

## 2. Applications of Machine Learning Techniques to Biomedicals

The field of Biomedicals encompasses a wide range of specialized areas such as genetics, pharmacology, and medical imaging, and is aimed at improving human health [12], [17], lessening the healthcare cost burden, and a trajectory toward personalized medicine. Biomedical problems can range from understanding the underlying causes of diseases to formulating testable hypotheses and developing new treatments and diagnostic tools. Machine learning techniques had in recent times played a vital role in solving biomedical problems by analyzing large and complex datasets that are difficult for humans to understand [18]. ML algorithms have been employed to identify underlying patterns in the data and make predictions about new, unseen data. This has helped researchers to validate their results, develop new drugs and improve patients' treatments. Additionally, ML technologies have been employed to identify potential drug targets, and patients' stratifications, and predict treatment outcomes. Some of the setbacks to using ML techniques in the Biomedical field are the rising issues of model complexities and lack of domain knowledge integration during and after the development of the data science modeling framework [10], [19]. Thus, model interpretation and explanation become crucial to create more trustworthy ML solutions that are useful to domain experts.

### 3. Interpretable Machine Learning Approaches in Biomedicals

Interpretable machine learning (IML) is a rapidly growing field that has the potential to revolutionize biomedical research [20]. IML allows researchers to analyze complex and high-dimensional data that is difficult for humans to understand, while also providing insights into the predictions of the model [21], [22]. Molnar [23], [24] highlighted that there are two types of IML interpretation approaches – intrinsic and post-hoc. The intrinsic approaches are incorporated into the model such as the global feature importance score, weights, decision tree models, etc, while the post-hoc model agnostic approaches that are employed after the choice of an ML model have been trained. Examples include SHAP, LIME, Partial dependence plots (PDP), etc. Also, the post-hoc approach can be global such as the building of a surrogate model to approximate a complex model or local approach for individual explanations such as LIME and SHAP techniques [25]–[27].

One of the key benefits of IML in biomedical research is the ability to identify the most important features of a certain disease or condition [28]. For example, IML techniques can be used to measure the importance of each feature in the data for the model's predictions, which can help researchers to understand which features are most important for a certain disease or condition. This can help researchers to validate the model's results, identify potential biases or errors, and develop new hypotheses. Figure 2 summarized the existing state-of-the-art IML technique as proposed by Azodi *et al.,* [16].

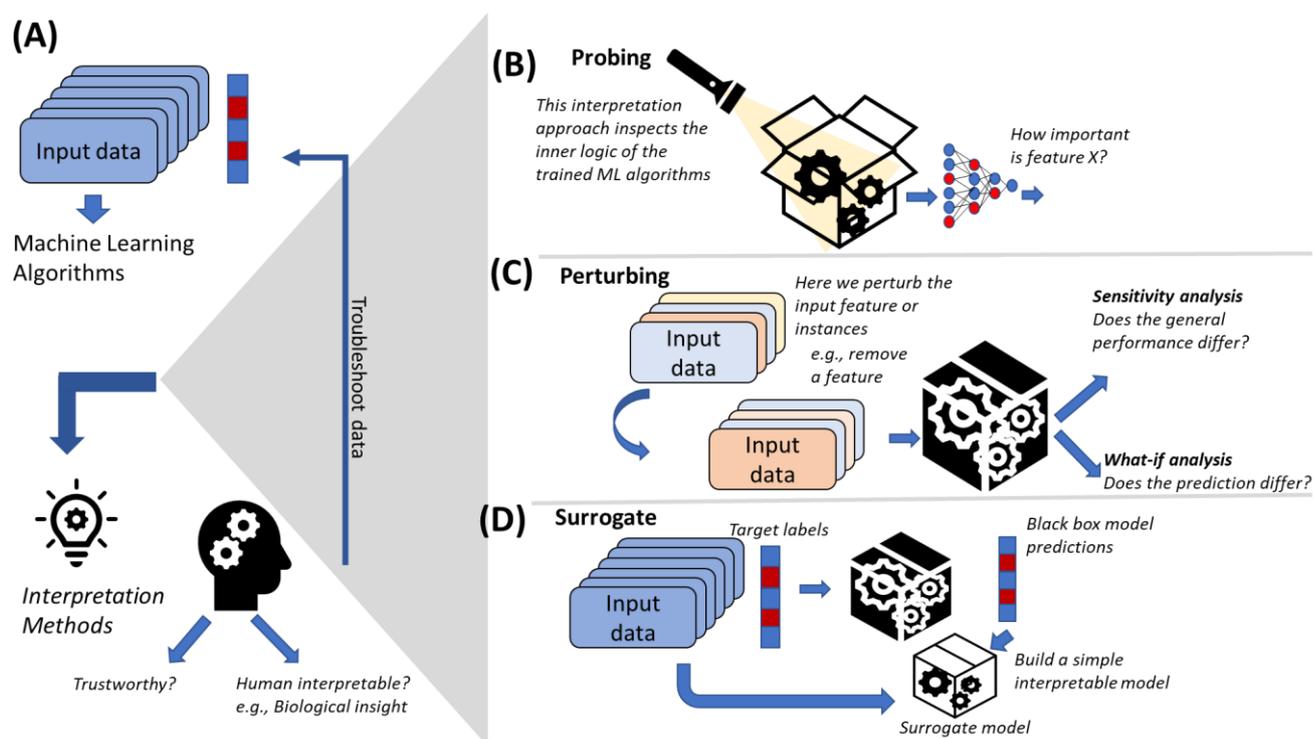

**Figure 2:** Interpretable Machine Learning procedure
- (A) Understanding the inner logic of an ML model (i.e., model interpretability) is important for troubleshooting during model training, generating biological insights, and instilling trust in the predictions made. (B) There are three main approaches to interpreting an ML model: probing, perturbing, and surrogates. Probing strategies involve inspecting the structure and parameters learned by a trained ML model (e.g., a DNN model pictured here) to better understand what features or combinations of features are important for driving the model's predictions. Perturbing strategies involve changing the values of one or more input features (e.g., setting all values to zero) and measuring the change in model performance (sensitivity analysis) or on the predicted label of a specific instance (what-if analysis). Finally, an easy IML model (e.g., linear regression or decision tree) can be trained to predict the predictions from ML models, acting as a surrogate.

Employing the use of IML techniques can provide local explanations for the predictions of an ML model [29]. This can be useful for understanding the predictions of an ML model for a specific sample or subgroup of samples. However, it's also important to understand the global behavior of the model, not just the predictions for individual samples [30]. For example, the feature importance weight score from decision tree models has been used as a global metric to understand the model behavior. In biomedical research, the choice of IML models is often used to analyze complex and high-dimensional data, such as genomic data, tabular data, and electronic health record data [31], [32] instead of novel approaches like the DNN [5], [7]. One of the reasons for this is model complexity and lack of confidence in the decisions of the black box models [33]. However, the choice of an IML model over high-performance black box models like DNN implies that the experts have to make crucial trade-offs such as sacrificing model accuracy with interpretability. Interpretable ML can help biomedical researchers to understand how a model is making its predictions and identify the features that are most important for the predictions [24]. Also, IML solutions can help to improve trust and acceptance of the model's results by stakeholders, including physicians, patients, and regulatory bodies [24], [32], [34].

Examples of IML algorithms include Linear Regression, Logistics Regression, and Decision Trees. Other notable ML models with lesser complexities frequently employed in the Biomedical domain include Naive Bayes, K-Nearest Neighbors, Random Forest, Support Vector Machines (SVMs), and XGBoost. Notable IML approaches include employing the use of Explainable Artificial Intelligence (XAI) models such as LIME, SHAP, and PDP, and Transparent models such as RuleFit, GAMs, and MARS. Some of the key interpretable metrics frequently used by experts to make valuable insights into the model's decisions include; **1) Feature importance:** This technique measures the importance of each feature in the data during model training. Features that have high importance are more likely to be relevant during the prediction phase. This metric can help researchers to understand which features are most important for a certain disease or condition. **2) Local interpretable model-agnostic explanations (LIME):** This technique generates an interpretable model that is like the original model but is simpler and can be used to explain the predictions of the original model locally. LIME can be used to understand the predictions of a model for a specific sample or subgroup of samples. **3) SHAP values:** This technique measures the contribution of each feature to the predictions of the model and can be used to understand how the model is using different features to make predictions. These techniques are commonly used in biomedical research to understand the predictions of ML models and identify the most important features of a certain disease or condition.

While IML can be very useful in biomedical research, some potential pitfalls [18], [32], [35] should be considered. Some of these include; **1) Over-simplification:** IML techniques can be used to simplify the predictions of a model, but this simplification can lead to a loss of accuracy or a reduction in the model's performance. Researchers should be careful not to over-simplify the model, as this can lead to inaccurate or unreliable results. **2) Limited scope:** Some IML techniques, such as LIME, are designed to provide local explanations for the predictions of a model [36]. This means that they can only be used to understand the predictions of the model for a specific sample or subgroup of samples. In biomedical research, it is important to understand the global behavior of the model and not just the predictions for individual samples. **3) Limited explanation:** IML techniques can be used to provide insight into the predictions of a model, but they may not be able to fully explain the underlying mechanisms or causes of a disease or condition. Researchers should be aware of the limitations of IML techniques and not rely solely on them to understand the underlying mechanisms of a disease or condition. **4) Data complexity:** Biomedical data often contains a high degree of complexity, heterogeneity, the curse of dimensionality, and noise, which makes it difficult to interpret the results of ML models [8]. The curse of dimensionality refers to the difficulty of accurately modeling high-dimensional data due to the exponential increase in the number of features and combinations of features that must be considered as the dimensionality of the data increases [11]. In the context of biomedical datasets, this can lead to overfitting, reduced model stability, and poor predictive performance, which can limit the utility of IML approaches. IML techniques can be used to understand the predictions of a model, but the complexity of the data can make it difficult to interpret the results [21]. **5) Model bias:** IML techniques can be used to identify potential biases in an ML model, but they may not be able to eliminate bias [27]. Researchers should be aware of the potential biases in the data and the model and use techniques such as data pre-processing and bias correction to mitigate bias. **6) Explainability vs Performance:** There is a trade-off between interpretability and performance. More interpretable models tend to be less complex and thus less accurate. Researchers should be aware of this trade-off and choose the appropriate level of interpretability for their specific use case.

To mitigate the limitations of IML, researchers have adopted various techniques. These include the use of IML libraries and packages such as ExplainerDashboard [37], AI Explainability 360 (AIX360), Alibi Explainer, ELI5, H20 ML resource, LIME, and SHAP values [25] to provide interpretable insights into model predictions without oversimplifying the model. Global interpretability methods (e.g., PDP, ALE) and counterfactual explanations have also been incorporated to understand the global behavior of trained ML models. Causal inferential methods, however, a lacking but some experts are incorporating statistical approaches to infer the underlying causal mechanisms that drive model predictions [38]–[40]. Dimensionality reduction methods (e.g., PCA, t-SNE) and feature selection methods have often been used to reduce data dimensionality and make it more interpretable [41], [42]. Data pre-processing and bias correction methods, as well as explainable artificial intelligence (XAI) methods, have also been used to mitigate biases in the data and model. Lastly, sensitivity analysis and cross-validation [42] are employed to assess the robustness of the model.

## 4. Enhancing Results of Machine Learning Models with Data Stratifications and Domain Knowledge Integration

A crucial aspect of concern that this paper stressed and has not been fully addressed in recent times is the issues relating to model stability in IML algorithms when training high-dimensional biomedical datasets as they can be prone to overfitting and poor performance. One approach to addressing the curse of dimensionality in IML is to use feature selection or dimensionality reduction techniques to identify the most informative features or reduce the overall dimensionality of the dataset [43]. However, these methods can also introduce additional complexity and potential for error, particularly if domain knowledge is not incorporated into the feature selection process. Another approach is to use ensemble methods or regularization techniques to improve the model stability and prevent overfitting. For example, using a combination of multiple models or incorporating penalties like LASSO, and ElasticNet for model complexity can help to improve performance and reduce the risk of overfitting [44], [45]. While the issue related to a lack of integrating domain knowledge analyses and interpretations in the development of the IML framework [13], [46] can be established by linking the IML results with the domain wealth of knowledge such as literature or databases. IML techniques can provide valuable insights into the predictions of ML models, but they may not be able to fully capture the complex and nuanced relationships between the features and the disease or condition being studied. Automatically linking the most ranked important features from trained IML models with potential domain knowledge databases can ease interpretations and implications of findings for domain experts.

In biomedical research, domain knowledge is critical for understanding the underlying mechanisms identified during and after the ML model training phase. For example, what are the implications of weighted scores (feature importance) identified for a cause of a disease or condition? Without the integration of domain knowledge expertise, it can be difficult to interpret the results of an IML model or validate the results accurately. For example, automatically linking the identified features' importance names to a disease/drug repository for plausible drug targets or therapy [47] could create insights and acceptability of the ML solutions by domain experts. Additionally, when analyzing genomic data, domain knowledge is essential to understand the functional and biological implications of specific genetic variations [48], [49]. Without it, researchers may not be able to fully interpret the results of an IML technique or validate the predictions of the ML model. Another example is when dealing with medical images, domain knowledge is essential to understand the characteristics of the images and the specific features that are relevant for the diagnosis or treatment of a disease. Without it, researchers may not be able to fully interpret the results of an IML model.

To improve the integration of problem data stratification and domain knowledge in IML frameworks, several steps can be taken. One effective approach is to split the problem dataset randomly into stratified k-fold CVs [50]. Each fold will consist of a training set and a testing set, and the stratification will be done such that there will be no data leakage. That is; the training set and testing set in each fold are completely antagonistic to each other. These k-fold CVs will then be trained using several pre-selected ML state-of-the-art algorithms. Feature importance from the best estimator models can then be aggregated for further domain knowledge analysis. To improve model performance and stability, the best-trained estimator parameters from each k-fold are further combined via an ensemble voting classifier and then used to retrain the k-fold CVs' most informative features identified. To integrate domain knowledge analyses in the IML framework, one can incorporate them into the model itself, such as linking domain-specific feature names or knowledge databases with the most informative features dimed important from the trained ML models. Using domain-specific evaluation metrics is also important for accurately evaluating the performance and interpretability of the IML model. The aggregated (pool of important features) feature-weighted score list from the CV folds can be further harnessed for downstream domain knowledge analyses and interpretations such as functional enrichment pathways and drug target repurposing [50]. Domain knowledge can be incorporated into IML models in several biomedical areas such as medical imaging, drug discovery, genomic data, medical diagnosis, and clinical decision support. Additional examples of domain knowledge approach that can be integrated into the IML framework include expert knowledge, patient/doctor focus group discussions, annotated medical images, protein structures, biological processes, genetic variations, disease features, or clinical protocols. Continuous learning from domain experts and updating the model with new knowledge is also important to improve accuracy, debug system, and interpretability.

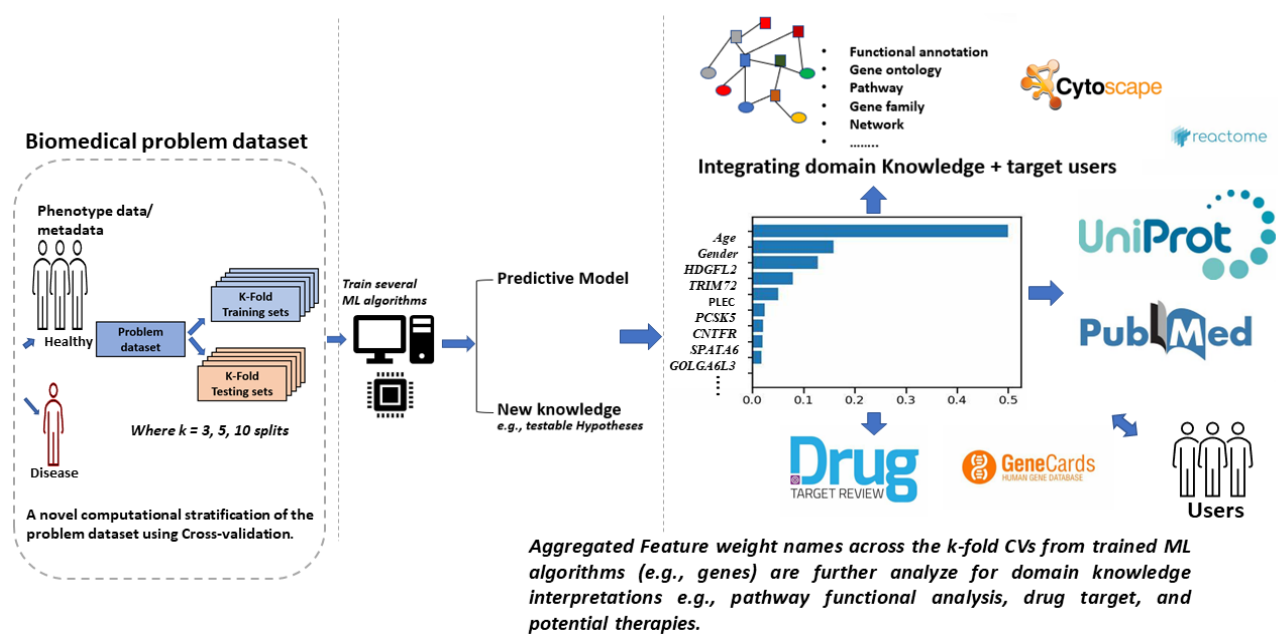

**Figure 4:** Proposed Interpretable ML framework for Biomedical Data Science Project.
An overview of interpretable ML framework integrating a computational splitting strategy of the problem dataset via a stratified k-fold cross-validation approach (i.e., with each fold having a training set and a testing set, and the splitting is done such that there is no information leakage between the training set and the testing set in each of the k-fold CVs). The introduction of cross-validation splits before training will improve model stability and generalizability abilities. Additionally, integrating the domain knowledge analyses and interpretation techniques to harness insights and knowledge extracted from ML-trained algorithms such as feature importance weights, saved model outcomes, etc. In the area of Bioinformatics such as genomic problems, the ML feature importance weighted score list can be names of genetic variants that can further be analyzed for domain knowledge interpretation e.g., pathway functional enrichment analysis, protein structure, drug targeting and repurposing, and potential therapies from databases such as UniProt [51], Cytoscape, Reactome [52], [53], GeneCards [54], and so on.

## 5. Conclusion

In conclusion, IML techniques can provide valuable insights into the predictions of ML models in biomedical research, but it is important to consider the potential pitfalls and limitations of these techniques. Researchers should be aware of the complexity of biomedical data, the trade-off between interpretability and performance, and the potential biases in the data and model. In collaboration with domain experts, data scientists can design, interpret, and validate IML results accurately. Thus, incorporating a computational splitting strategy of the problem dataset and domain knowledge analyses and interpretations into the existing IML framework could go a long way in addressing crucial issues and creating more robust and trustworthy ML solutions in the Biomedical areas of applications.

**Competing interests:** The authors declare no competing interests.

## References

[1] "What is Machine Learning? | IBM." https://www.ibm.com/cloud/learn/machine-learning (accessed Oct. 07, 2022).

[2] B. Liu, "Supervised Learning," *Web Data Mining*, pp. 63–132, 2011, doi: 10.1007/978-3-642-19460-3_3.

[3] P. Cunningham, M. Cord, and S. J. Delany, "Supervised learning," *Cognitive Technologies*, pp. 21–49, 2008, doi: 10.1007/978-3-540-75171-7_2/COVER.

[4] P. S. Bernard *et al.*, "Supervised risk predictor of breast cancer based on intrinsic subtypes," *Journal of Clinical Oncology*, vol. 27, no. 8, pp. 1160–1167, Mar. 2009, doi: 10.1200/JCO.2008.18.1370.

[5] P. Baldi, "Deep Learning in Biomedical Data Science," *https://doi.org/10.1146/annurev-biodatasci-080917-013343*, vol. 1, no. 1, pp. 181–205, Jul. 2018, doi: 10.1146/ANNUREV-BIODATASCI-080917-013343.

[6] A. Voulodimos, N. Doulamis, A. Doulamis, and E. Protopapadakis, "Deep Learning for Computer Vision: A Brief Review," *Comput Intell Neurosci*, vol. 2018, 2018, doi: 10.1155/2018/7068349.

[7] G. Ras, M. van Gerven, and P. Haselager, "Explanation Methods in Deep Learning: Users, Values, Concerns and Challenges," pp. 19–36, 2018, doi: 10.1007/978-3-319-98131-4_2/COVER.

[8] R. Poyiadzi, X. Renard, T. Laugel, R. Santos-Rodriguez, and M. Detyniecki, "Understanding surrogate explanations: the interplay between complexity, fidelity and coverage".

[9] "Biomedical Data Repositories and Knowledgebases | Data Science at NIH." https://datascience.nih.gov/data-ecosystem/biomedical-data-repositories-and-knowledgebases (accessed Oct. 06, 2022).

[10] R. Bellazzi, M. Diomidous, I. N. Sarkar, K. Takabayashi, A. Ziegler, and A. T. McCray, "Data analysis and data Mining: Current issues in biomedical informatics," *Methods Inf Med*, vol. 50, no. 6, pp. 536–544, 2011, doi: 10.3414/ME11-06-0002/ID/JR0002-17.

[11] C. Xu and S. A. Jackson, "Machine learning and complex biological data," *Genome Biol*, vol. 20, no. 1, pp. 1–4, Apr. 2019, doi: 10.1186/S13059-019-1689-0/FIGURES/2.

[12] Kording, K.P., Benjamin, A., Farhoodi, R. and Glaser, J.I., 2018, January. The roles of machine learning in biomedical science. In *Frontiers of Engineering: Reports on Leading-Edge Engineering from the 2017 Symposium*. National Academies Press.

[13] P. Kaur, A. Singh, and I. Chana, "Computational Techniques and Tools for Omics Data Analysis: State-of-the-Art, Challenges, and Future Directions," *Archives of Computational Methods in Engineering 2021 28:7*, vol. 28, no. 7, pp. 4595–4631, Feb. 2021, doi: 10.1007/S11831-021-09547-0.

[14] S. Yadav and S. Shukla, "Analysis of k-Fold Cross-Validation over Hold-Out Validation on Colossal Datasets for Quality Classification," *Proceedings - 6th International Advanced Computing Conference, IACC 2016*, pp. 78–83, Aug. 2016, doi: 10.1109/IACC.2016.25.

[15] C. Su, J. Tong, Y. Zhu, P. Cui, and F. Wang, "Network embedding in biomedical data science," *Brief Bioinform*, vol. 21, no. 1, pp. 182–197, Jan. 2020, doi: 10.1093/BIB/BBY117.

[16] C. B. Azodi, J. Tang, and S. H. Shiu, "Opening the Black Box: Interpretable Machine Learning for Geneticists," *Trends in Genetics*, vol. 36, no. 6, pp. 442–455, Jun. 2020, doi: 10.1016/J.TIG.2020.03.005.

[17] Y. Peng, Z. Wu, and J. Jiang, "A novel feature selection approach for biomedical data classification," *J Biomed Inform*, vol. 43, no. 1, pp. 15–23, Feb. 2010, doi: 10.1016/J.JBI.2009.07.008.

[18] H. Han and X. Liu, "The challenges of explainable AI in biomedical data science," *BMC Bioinformatics 2021 22:12*, vol. 22, no. 12, pp. 1–3, Jan. 2022, doi: 10.1186/S12859-021-04368-1.

[19] A. Holzinger and I. Jurisica, "Knowledge discovery and data mining in biomedical informatics: The future is in integrative, interactive machine learning solutions," *Lecture Notes in Computer Science (including subseries Lecture Notes in Artificial Intelligence and Lecture Notes in Bioinformatics)*, vol. 8401, pp. 1–18, 2014, doi: 10.1007/978-3-662-43968-5_1/COVER.

[20] "Interpretable Machine Learning." https://christophm.github.io/interpretable-ml-book/ (accessed Oct. 26, 2022).

[21] A. Vellido, J. D. Martín-Guerrero, and P. J. G. Lisboa, "Making machine learning models interpretable", Accessed: Oct. 08, 2022. [Online]. Available: http://www.i6doc.com/en/livre/?GCOI=28001100967420.

[22] M. Du, N. Liu, and X. Hu, "Techniques for interpretable machine learning," *Commun ACM*, vol. 63, no. 1, pp. 68–77, Jan. 2020, doi: 10.1145/3359786.

[23] Molnar, C., 2020. *Interpretable machine learning*. Lulu. com.

[24] C. Molnar, G. Casalicchio, and B. Bischl, "Interpretable Machine Learning – A Brief History, State-of-the-Art and Challenges," *Communications in Computer and Information Science*, vol. 1323, pp. 417–431, 2020, doi: 10.1007/978-3-030-65965-3_28/FIGURES/2.

[25] Y. Nohara, S. Kumamoto, H. Kumamoto, H. Soejima, and J. N. Nakashima, "Explanation of Machine Learning Models Using Improved Shapley Additive Explanation," pp. 546–546, Sep. 2019, doi: 10.1145/3307339.3343255.

[26] C. Molnar, G. Casalicchio, and B. Bischl, "iml: An R package for Interpretable Machine Learning Software • Review • Repository • Archive", doi: 10.21105/joss.00786.

[27] C. A. Scholbeck, C. Molnar, C. Heumann, B. Bischl, and G. Casalicchio, "Sampling, Intervention, prediction, aggregation: A generalized framework for model-agnostic interpretations," *Communications in Computer and Information Science*, vol. 1167 CCIS, pp. 205–216, 2020, doi: 10.1007/978-3-030-43823-4_18/FIGURES/2.

[28] Y. R. Cho and M. Kang, "Interpretable machine learning in bioinformatics," *Methods*, vol. 179, pp. 1–2, Jul. 2020, doi: 10.1016/J.YMETH.2020.05.024.

[29] W. J. Murdoch, C. Singh, K. Kumbier, R. Abbasi-Asl, and B. Yu, "Interpretable machine learning: definitions, methods, and applications," *Proc Natl Acad Sci U S A*, vol. 116, no. 44, pp. 22071–22080, Jan. 2019, doi: 10.1073/pnas.1900654116.

[30] O. Sagi and L. Rokach, "Approximating XGBoost with an interpretable decision tree," *Inf Sci (N Y)*, vol. 572, pp. 522–542, Sep. 2021, doi: 10.1016/J.INS.2021.05.055.

[31] M. A. Ahmad, C. Eckert, and A. Teredesai, "Interpretable Machine Learning in Healthcare," *Proceedings of the 2018 ACM International Conference on Bioinformatics, Computational Biology, and Health Informatics*, vol. 21, pp. 559–560, Aug. 2018, doi: 10.1145/3233547.

[32] T. Abdullah, M. Zahid, W. A.- Symmetry, and undefined 2021, "A review of interpretable ml in healthcare: Taxonomy, applications, challenges, and future directions," *mdpi.com*, 2021, doi: 10.3390/sym13122439.

[33] C. Rudin, "Stop explaining black box machine learning models for high stakes decisions and use interpretable models instead," *Nature Machine Intelligence 2019 1:5*, vol. 1, no. 5, pp. 206–215, May 2019, doi: 10.1038/s42256-019-0048-x.

[34] U. Kamath and J. Liu, "Post-Hoc Interpretability and Explanations," *Explainable Artificial Intelligence: An Introduction to Interpretable Machine Learning*, pp. 167–216, 2021, doi: 10.1007/978-3-030-83356-5_5.

[35] M. Ghassemi, T. Naumann, P. Schulam, A. L. Beam, I. Y. Chen, and R. Ranganath, "A Review of Challenges and Opportunities in Machine Learning for Health," *AMIA Summits on Translational Science Proceedings*, vol. 2020, p. 191, 2020, Accessed: Oct. 07, 2022. [Online]. Available: /pmc/articles/PMC7233077/

[36] "marcotcr/lime: Lime: Explaining the predictions of any machine learning classifier." https://github.com/marcotcr/lime (accessed Nov. 03, 2022).

[37] "explainerdashboard · PyPI." https://pypi.org/project/explainerdashboard/ (accessed Oct. 07, 2022).

[38] B. C. Saul, M. G. Hudgens, and M. E. Halloran, "Causal Inference in the Study of Infectious Disease," *Handbook of Statistics*, vol. 36, pp. 229–246, Jan. 2017, doi: 10.1016/BS.HOST.2017.07.002.

[39] S. Weichwald, T. Meyer, O. Özdenizci, B. Schölkopf, T. Ball, and M. Grosse-Wentrup, "Causal interpretation rules for encoding and decoding models in neuroimaging," *Neuroimage*, vol. 110, pp. 48–59, Apr. 2015, doi: 10.1016/J.NEUROIMAGE.2015.01.036.

[40] Imbens, G.W. and Rubin, D.B., 2015. *Causal inference in statistics, social, and biomedical sciences*. Cambridge University Press.

[41] P. Dayan, "Unsupervised Learning".

[42] N. A. Diamantidis, D. Karlis, and E. A. Giakoumakis, "Unsupervised stratification of cross-validation for accuracy estimation," *Artif Intell*, vol. 116, no. 1–2, pp. 1–16, Jan. 2000, doi: 10.1016/S0004-3702(99)00094-6.

[43] F. Model, P. Adorján, A. Olek, and C. Piepenbrock, "Feature selection for DNA methylation based cancer classification," *Bioinformatics*, vol. 17, no. suppl_1, pp. S157–S164, Jun. 2001, doi: 10.1093/BIOINFORMATICS/17.SUPPL_1.S157.

[44] O. Sagi and L. Rokach, "Ensemble learning: A survey," *Wiley Interdiscip Rev Data Min Knowl Discov*, vol. 8, no. 4, p. e1249, Jul. 2018, doi: 10.1002/WIDM.1249.

[45] X. Dong, Z. Yu, W. Cao, Y. Shi, and Q. Ma, "A survey on ensemble learning," *Frontiers of Computer Science 2019 14:2*, vol. 14, no. 2, pp. 241–258, Aug. 2019, doi: 10.1007/S11704-019-8208-Z.

[46] S. J. Sammut *et al.*, "Multi-omic machine learning predictor of breast cancer therapy response," *Nature 2021 601:7894*, vol. 601, no. 7894, pp. 623–629, Dec. 2021, doi: 10.1038/s41586-021-04278-5.

[47] J. Reimand *et al.*, "Pathway enrichment analysis and visualization of omics data using g:Profiler, GSEA, Cytoscape and EnrichmentMap," *Nature Protocols 2019 14:2*, vol. 14, no. 2, pp. 482–517, Jan. 2019, doi: 10.1038/s41596-018-0103-9.

[48] M. Ghoussaini *et al.*, "Open Targets Genetics: systematic identification of trait-associated genes using large-scale genetics and functional genomics," *Nucleic Acids Res*, vol. 49, no. D1, pp. D1311–D1320, Jan. 2021, doi: 10.1093/NAR/GKAA840.

[49] H. Tang and P. D. Thomas, "Tools for Predicting the Functional Impact of Nonsynonymous Genetic Variation," *Genetics*, vol. 203, no. 2, pp. 635–647, Jun. 2016, doi: 10.1534/GENETICS.116.190033.

[50] A. Onoja *et al.*, "An explainable model of host genetic interactions linked to COVID-19 severity," *Communications Biology 2022 5:1*, vol. 5, no. 1, pp. 1–14, Oct. 2022, doi: 10.1038/s42003-022-04073-6.

[51] "UniProt." https://www.uniprot.org/ (accessed Feb. 10, 2023).

[52] G. Wu, E. Dawson, A. Duong, R. Haw, and L. Stein, "ReactomeFIViz: a Cytoscape app for pathway and network-based data analysis," *F1000Res*, vol. 3, Sep. 2014, doi: 10.12688/F1000RESEARCH.4431.2.

[53] B. Jassal *et al.*, "The reactome pathway knowledgebase," *Nucleic Acids Res*, vol. 48, no. D1, pp. D498–D503, Jan. 2020, doi: 10.1093/NAR/GKZ1031.

[54] "GeneCards - Human Genes | Gene Database | Gene Search." https://www.genecards.org/ (accessed Feb. 10, 2023).